\documentclass{article}

\usepackage{PRIMEarxiv}

\usepackage[utf8]{inputenc} 
\usepackage[T1]{fontenc}    
\usepackage{hyperref}       
\usepackage{url}            
\usepackage{booktabs}       
\usepackage{amsfonts}       
\usepackage{nicefrac}       
\usepackage{microtype}      
\usepackage{lipsum}
\usepackage{fancyhdr}       
\usepackage{graphicx}       
\graphicspath{{media/}}     

\usepackage[numbers]{natbib}

\usepackage{multirow}
\usepackage{float}
\usepackage{subfig} 
\usepackage{amsmath}
\usepackage{mathtools}
\usepackage{bm}
\usepackage{multirow}
\usepackage{enumitem}
\usepackage[ruled,vlined,linesnumbered,noend]{algorithm2e}
\usepackage[table,xcdraw]{xcolor}

\usepackage{combelow}

\usepackage{xcolor}

\usepackage[T2A,T1]{fontenc}
\DeclareTextSymbolDefault{\cyre}{T2A}
\DeclareTextSymbolDefault{\CYRE}{T2A}
\DeclareTextSymbolDefault{\cyrf}{T2A}
\DeclareTextSymbolDefault{\cyra}{T2A}
\DeclareTextSymbolDefault{\cyrn}{T2A}
\DeclareTextSymbolDefault{\cyrr}{T2A}
\DeclareTextSymbolDefault{\cyri}{T2A}
\DeclareTextSymbolDefault{\cyrm}{T2A}
\DeclareTextSymbolDefault{\cyrv}{T2A}
\DeclareTextSymbolDefault{\cyrz}{T2A}
\DeclareTextSymbolDefault{\cyrl}{T2A}
\DeclareTextSymbolDefault{\cyru}{T2A}
\DeclareTextSymbolDefault{\cyrt}{T2A}
\DeclareTextSymbolDefault{\cyrp}{T2A}
\DeclareTextSymbolDefault{\cyro}{T2A}
\DeclareTextSymbolDefault{\cyrh}{T2A}
\DeclareTextSymbolDefault{\cyrs}{T2A}
\DeclareTextSymbolDefault{\cyrk}{T2A}
\DeclareTextSymbolDefault{\cyrb}{T2A}
\DeclareTextSymbolDefault{\cyshch}{T2A}
\DeclareTextSymbolDefault{\CYRP}{T2A}
\DeclareTextSymbolDefault{\cyrshch}{T2A}
\DeclareTextSymbolDefault{\cyrya}{T2A}
\DeclareTextSymbolDefault{\cyrch}{T2A}
\DeclareTextSymbolDefault{\cyrg}{T2A}
\DeclareTextSymbolDefault{\CYRN}{T2A}
\DeclareTextSymbolDefault{\cyrhrdsn}{T2A}
\DeclareTextSymbolDefault{\CYRI}{T2A}
\DeclareTextSymbolDefault{\cyrsh}{T2A}
\DeclareTextSymbolDefault{\cyrd}{T2A}
\DeclareTextSymbolDefault{\cyrje}{T2A}
\DeclareTextSymbolDefault{\CYRK}{T2A}
\DeclareTextSymbolDefault{\cyrc}{T2A}
\DeclareTextSymbolDefault{\cyrzh}{T2A}

\title{
Semantic Change Detection for the Romanian Language
}

\author{
  Ciprian-Octavian Truic{\u{a}}, Victor Tudose, and Elena-Simona Apostol\\
  University Politehnica of Bucharest, Bucharest, Romania \\
  \texttt{
    ciprian.truica@upb.ro, 
    victor.tudose@stud.acs.upb.ro, 
    elena.apostol@upb.ro
    }
}

\begin{document}
\maketitle

\begin{abstract}
Automatic semantic change methods try to identify the changes that appear over time in the meaning of words by analyzing their usage in diachronic corpora. 
In this paper, we analyze different strategies to create static and contextual word embedding models, i.e., Word2Vec and ELMo, on real-world English and Romanian datasets.
To test our pipeline and determine the performance of our models, we first evaluate both word embedding models on an English dataset (SEMEVAL-CCOHA).
Afterward, we focus our experiments on a Romanian dataset, and we underline different aspects of semantic changes in this low-resource language, such as meaning acquisition and loss.
The experimental results show that, depending on the corpus, the most important factors to consider are the choice of model and the distance to calculate a score for detecting semantic change.
\end{abstract}


\maketitle

\section{Introduction}\label{sec:introduction}

Language is in a continuous process of change that occurs permanently, language change being the phenomenon that drives language evolution, as a process of adaptation to the environment and the ways other speakers use the language~\cite{Armaselu2022a,Armaselu2022b}.
The various instances of language change are classified into different categories, such as regular phonetic changes, changes in word usage, and changes in the way words appear together, i.e., syntactic changes. 
Although it is usually a continuous process that follows regular patterns, very abrupt changes in the meanings of words can still occur, usually motivated by a change in the context a community lives in~\cite{Hamilton2016,Dubossarsky2017,Chiru2021}.

Semantic change, as a phenomenon permanently present in language evolution, is an important aspect that should be taken into account when working with historical data\cite{Armaselu2021}.
Historical linguists, lexical typologists, and other humanities and social science experts have studied the meaning of words and how it changes over time.
The method used by these experts is known as "close reading", being the manual study of texts.

The growing availability of digitalized historical corpora has enabled larger-scale quantitative studies of language evolution and a rise in the usage of computational approaches, both automatic and semi-automatic, for computational linguistics tasks.
New approaches to semantic change detection had appeared with the recent advances in Natural Language Processing such as word embedding techniques.
The main approaches use either non-contextualized (static) word embeddings that generate one embedding per word regardless of the context~\cite{Gong2020,Bizzoni2019,Tsakalidis2020} and contextualized word embeddings that generate multiple contested dependent embeddings for the same word~\cite{Rodina2020,Kanjirangat2020}.

The main objective of this work is to analyze the semantic change in Romanian.
To achieve this, we:\\
(1) develop an architecture for determining semantic change using two parallel corpora from different time periods to model static and contextual word embeddings;\\
(2) determine the accuracy of our model using two lists of words, one with words that undergo semantic change and one with words for which their sense remains stable over the time periods;\\
(3) explore how different metrics can be affected by the corpora employed in our analysis;\\
(4) create a web application to visualize and analyze the models.

The rest of this paper is structured as follows:
In Section~\ref{sec:relatedwork}, we describe some of the most relevant current similar works.
Section~\ref{sec:solution} introduces and presents our proposed solution.
In Section~\ref{sec:results}, we briefly describe our datasets and experimental setup and analyze the results obtained in our experiments.
Section~\ref{sec:discussions} provides an in-depth discussion of our results.
Section~\ref{sec:conclusions} concludes our paper and hints at future research.

\section{Related Work}~\label{sec:relatedwork}

In the current literature, static and contextual word embedding models together with different distance metrics have been used to determine semantic change.
The main idea is to create word embeddings for parallel corpora that span over different periods of time. 
Afterward, a score is calculated based on the distance between two vector representations from different periods of time of the same word.
If the score is above a given threshold, then the word undergoes a semantic change.

Gong et al. (2020)~\cite{Gong2020} create condition-specific embeddings that consider spatial-temporal dimensions to model the words in a corpus and capture language evolution across both time and location.
Bizzoni et al. (2019)~\cite{Bizzoni2019} use hyperbolic embeddings~\cite{Nickel2017} to detect semantic changes that occur in domain-specific literature.
Tsakalidis and Liakata (2020)~\cite{Tsakalidis2020} use deep neural networks, e.g., sequence-to-sequence (Seq2Seq) models, together with Word2Vec Skip-Gram model~\cite{Mikolov2013} to create new word representation. Semantic change is evaluated by  calculating the average cosine similarity of word representations for different time periods.
Wegmann et al. (2020)~\cite{Wegmann2020} use local neighborhood~\cite{Hamilton2016} to determine semantic shifts using static word embeddings.

Recent studies have also explored the use of contextual word embedding and transformer models, e.g., ELMo (Embeddings from Language Models)~\cite{Peters2018} and BERT (Bidirectional Encoder Representations from Transformers)~\cite{Devlin2019}, in lexical semantic change tasks.
These studies utilized contextualized word representations to measure semantic shifts of words over time and evaluated their approach on large diachronic English corpora as well as other high-resource languages~\cite{Giulianelli2020,Kanjirangat2020,Rodina2020,Basile2020}, e.g., Latin, Italian, German, and Swedish. The results demonstrated that their proposed models captured a range of synchronic and diachronic linguistic aspects, and outperformed baselines based on normalized frequency difference or cosine distance methods.

\section{Proposed Solution}~\label{sec:solution}

\begin{figure}[!ht]
\centering
\includegraphics[width=0.8\columnwidth]{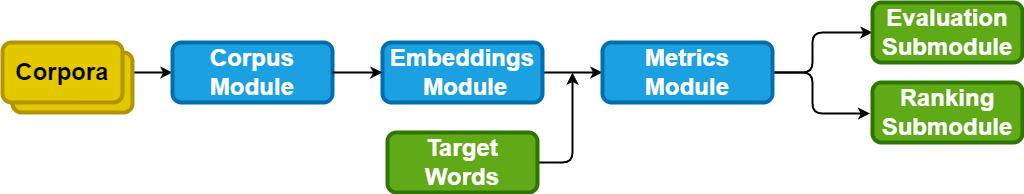}
\caption{The Proposed Architecture }
\label{fig:arch}
\end{figure}

Figure~\ref{fig:arch} presents our proposed architecture consisting of three main modules, i.e., Corpus, Embeddings, and Metric Modules. 
Our solution is designed to facilitate training multiple models on different corpora.
After the model is trained, we store the results obtained by each evaluation metric as well as the ranked list of the words that undergo semantic changes.
We implement our semantic change detection architecture using Python v3.7.
The code is available on GitHub at \url{https://github.com/DS4AI-UPB/SemanticChange-RO}.

\paragraph{Corpus Module}
This module performs text preprocessing. 
Given a corpus of documents, we employ the following steps.
(1) \textbf{Vocabulary extraction}: used by the Embeddings Module to extract the terms present in the corpora; and 
(2) \textbf{Time Interval Merging}: this takes a set of texts from different moments in time and a list of intervals, and merges the texts into new corpora that are obtained by merging the initial texts according to interval adherence.

\paragraph{Embeddings Module}
This module receives a set of time interval split corpora and the vocabulary.
For each corpus, we train static or contextual word embedding models, i.e., Word2Vec and ELMO, respectively.

\textit{Static Word Embeddings. }
We choose Word2Vec Skip-Gram with Negative Sampling (SGNS) as the static word embedding model based on the results of Hamilton et al. (2016b)~\cite{Hamilton2016-diachronic}.
Table~\ref{tab:word2vec_sgns_params} presents the parameters used for training the vectors.

\begin{table}[!htbp]
    \centering
    \caption{Word2Vec SGNS model parameters}
    \label{tab:word2vec_sgns_params} 
    \begin{tabular}{ |l|l|r|  }
         \hline
         \multicolumn{3}{|c|}{\textbf{Word2Vec SGNS Parameters}} \\ 
         \hline
         \textbf{Name} & \textbf{Description} & \textbf{Value}\\
         \hline
         min count       &minimum number of word occurrences &   1\\
         vector size  & dimensionality of vectors   &100\\
         window & max distance between current and predicted word&  5\\
         alpha & learning rate&  0.025\\
         negative  & how many negative samples are used &5\\
         ns exponent & correlation between frequency and negative sampling   &1\\
         epochs  & number of training iterations &5\\
         \hline
\end{tabular}
\end{table}

\begin{figure}[!h]
\centering
\includegraphics[width=0.75\columnwidth]{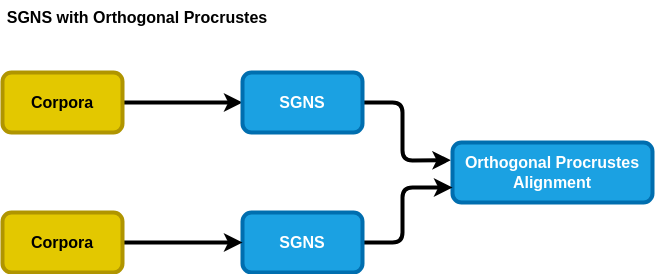}
\caption{SGNS-OP Architecture}
\label{fig:sgnsop} 
\end{figure}

We use two strategies to train our word embeddings.
The first implementation, i.e., SGNS-OP, trains the vectors on the time interval split corpora in parallel and uses the orthogonal Procrustes method to align the word embeddings (Figure \ref{fig:sgnsop}).
The Procrustes method~\cite{Schonemann1966} states that given two matrices $A$ and $B$ we need to find a matrix $\Omega$  such as $R = {argmin}_{\Omega}( \mid  \mid  \Omega \cdot A - B  \mid  \mid _{F})$, where 
$\Omega^T \cdot \Omega = I$ and $ \mid  \mid A \mid  \mid _F = \sqrt{\sum^n_{i,j = 1}  \mid a_{i,j} \mid ^2}$ called also the matrix norm or the Frobenius norm.

\begin{figure}[!h]
\centering
\includegraphics[width=0.75\columnwidth]{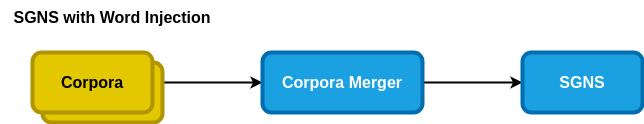}
\caption{SGNS-WI Architecture}
\label{fig:sgnswi}
\end{figure}

The second strategy for training the static word embeddings, i.e., SGNS-WI, makes use of word injection.
Using this method, we train the word embeddings on all the corpora at once but we tag some target words to obtain two different embeddings that correspond to the corpus the word originates from (Figure~\ref{fig:sgnswi}).

In our implementation, we use \href{https://radimrehurek.com/gensim/}{Gensim} \cite{Rehurek2010} to train both static word embeddings, i.e., SGNS-OP and SGNS-WI, respectively.

\textit{Contextual Word Embeddings.}
For the contextual word embeddings, we use \href{https://allenai.org/allennlp}{AllenNLP}~\cite{Gardner2018} ELMo~\cite{Peters2018} implementation.
One reason to use contextualized embeddings in the case of semantic change detection is to better model polysemy.
Since the word's meaning is disambiguated by its surrounding, naturally, a way of encoding the immediate surrounding of a word could help with capturing the particular sense a word has in its respective context.
The implementation of ELMo uses several modules.
(1) \textbf{Dataset Reader Module}: A module for preparing the corpus that (a) extracts the tokens, and (b) creates a token index.
(2) \textbf{Uncontextualised Embedding Module}: A module that creates the non-contextualized embeddings used by ELMo. This module uses a character-level Convolutional Neural Network that creates character embeddings of size 16. The character embeddings are then passed to a second layer that outputs non-contextualized word embeddings with a dimensionality of 256.
(3) \textbf{Contextualizer}: A contextualizing layer that extracts the context for the  embeddings. We used a BiLSTM contextualizer which concatenates a forward contextualized embedding to a backward contextualized embedding.
(4) \textbf{Bidirectional Language Model Loss Module}: Computes the loss for the ELMo model based on the sum of the log probabilities of all the sentences in the corpus.

\begin{figure}[!h]
\centering
\includegraphics[width=0.75\columnwidth]{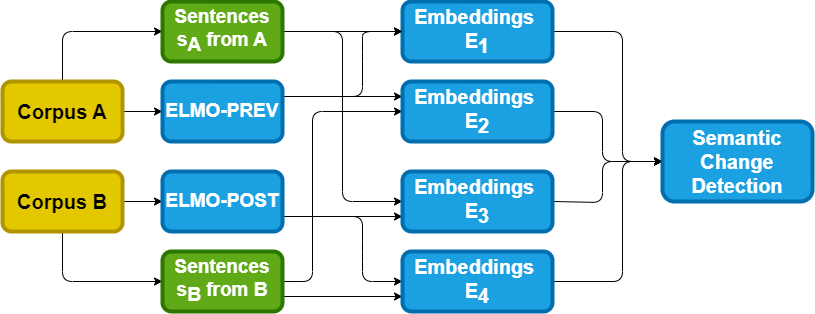}
\caption{Contextual Models Architecture}
\label{fig:elmo}
\end{figure}

We create a contextual word embedding model for each time window corpus.
To analyze the semantic change, for each corpus used to create a model we take all the sentences that contain a target word and then extract the embeddings from these sentences using the ELMo models trained on each corpus.
Thus, given two corpora $A$ and $B$, we train a contextual word embedding mode for each: 
(1) ELMO-PREV trained on $A$, and
(2) ELMO-POST trained on $B$.
Using these embeddings we create 4 sentence embeddings, as follows:
1) $\bm{E_1}$ are the embeddings obtained when applying the ELMO-PREV on the sentences $s_A \in A$ 
2) $\bm{E_2}$ are the embeddings obtained when applying the ELMO-PREV on the sentences $s_B \in B$
3) $\bm{E_3}$ are the embeddings obtained when applying the ELMO-POST on the sentences $s_A \in A$
4) $\bm{E_4}$ are the embeddings obtained when applying the ELMO-POST on the sentences $s_b \in A$
After obtaining these embeddings, we use different metrics to compare them and detect semantic changes.
Figure~\ref{fig:elmo} presents the architecture to train the ELMO-PREVand ELMO-POST models.

\paragraph{Metrics Module}
We employ multiple metrics to detect different nuances of semantic change that words undergo over time~\cite{Hamilton2016,Rubenstein1965,Wegmann2020}, e.g., local neighborhoods metrics determine better cultural changes, while global displacement measurements tend to detect regular semantic changes.
Thus, the metrics module receives the word embeddings and the target words for testing the semantic change and then computes a distance function to determine if the word's meaning undergoes any changes over time.

For every corpus, we create associated lists of target words to test the models' accuracy with words that undergo semantic change and words that kept their sense.
We use the evaluation metrics with these lists to determine if we manage to detect correctly semantic changes by determining if the word maintains its ranking, thus, its meaning.
Furthermore, when testing each model's performance using the lists of target words, we also record the number of true positives (TP), true negatives (TN), false positives (FP), and false negatives (FN).
The evaluation metrics and the rankings are stored using Evaluation Submodule and Ranking Submodule, respectively.

\textit{Static Word Embeddings Metrics.}
For the static word embedding models, given two $n$-dimensional static word embeddings $\bm{a} \in \mathbb{R}^n$ and $\bm{b} \in \mathbb{R}^n$ for the same word $w$, we use the following distances:

\textbf{Euclidean distance} computes the global displacement between the embeddings, indicating a change in the semantic properties: $d(\bm{a},\bm{b})=\sqrt{\sum_{i=1}^{n}(\bm{a}_{i}-\bm{b}_{i})^2}$

\textbf{Manhattan distance} measures the global distance between two embeddings taking into account the dimensionality and number of vector components: $d(\bm{a},\bm{b})=\sum_{i=1}^{n} \mid \bm{a}_i - \bm{b}_i \mid$.

\textbf{Canberra distance} is a weighted measure of the Manhatten distance that proved it provides good results when used for detecting semantic change~\cite{Basile2020,Giulianelli2020} 
$d(\bm{a},\bm{b}) = \sum_{i = 1}^{n} \frac{ \mid \bm{a}_i - \bm{b}_i \mid }{ \mid \bm{a}_i \mid + \mid \bm{b}_i \mid }$

\textbf{Cosine distance} determines the dissimilarity between the embeddings to detect semantic change: 
$d(\bm{a},\bm{b})= 1 - \frac{\sum_{i=1}^{n}\bm{a}_i \cdot \bm{b}_i}{\sqrt{\sum_{i=1}^{n}\bm{a}_i^2}\sqrt{\sum_{i=1}^{n} \bm{b}_i^2}}$ 
    
\textbf{Bray-Curtis distance} measures the dissimilarity between two word embeddings that considers their values relative magnitudes:
$d(\bm{a},\bm{b})) = \sum_{i = 1}^{n} \frac{ \mid \bm{a}_i - \bm{b}_i \mid }{ \mid \bm{a}_i + \bm{b}_i \mid }$.

\textbf{Correlation distance}: computes the  correlation between two word embeddings:
    $d(\bm{a},\bm{b})) = 1-\frac{(\bm{a}-\bar{\bm{a}}) \cdot (\bm{b} - \bar{\bm{b}})}{ \mid  \mid \bm{a} - \bar{\bm{a}} \mid  \mid _2  \mid  \mid \bm{b} - \bar{\bm{b}} \mid  \mid _2}$, where $\bar{\bm{a}}$ and $\bar{\bm{b}}$ are the mean of the elements of $\bm{a}$ and $\bm{b}$, respectively, and $\cdot$ is used to denote the dot product between two vectors.

\textit{Contextual Word Embeddings Metrics.}
For measuring the performance of the contextual word embeddings models for the semantic change detection task, we need to consider that we are analyzing multiple word embeddings for the same word in the same period of time in that are context dependent.
Thus, we must consider different set-based methods that group together the word embeddings, i.e., Average Pairwise Distances, Jensen-Shannon divergence, and Cluster Count based on Affinity Propagation.

\textbf{Average Pairwise Distances (APD)} are a class of distance functions  that compute a given distance $d(\cdot)$ for every possible pair of vectors $<\bm{a}, \bm{b}>$ from two sets of vectors $A$ and $B$~\cite{Giulianelli2020}: $APD(A,B,d)=\frac{1}{ \mid A \mid \cdot  \mid B \mid }\sum_{ \bm{a} \in A, \bm{b} \in B} \mid d(\bm{a}, \bm{b}) \mid$. We compute the Average Pairwise Distances using the following distances: Euclidean, Manhattan, Canberra, and Cosine. 

\textbf{Jensen-Shannon divergence} is a method for measuring similarity between two probability distributions $P$ and $Q$ based on Kullback-Leibler divergence that always outputs a finite value: $JSD(P,Q)= \frac{1}{2}(D(P,M) + D(M,Q))$, where $M$ is the average of the two probability distributions $M=\frac{1}{2}(P+Q)$ and  $D(A, B)$ is the Kullback-Leibler divergence.

\textbf{Cluster Count} obtains similar vector clusters by using the Affinity Propagation algorithm~\cite{Frey2007}.

\begin{figure}[!htbp]
\centering
\includegraphics[width=0.75\columnwidth]{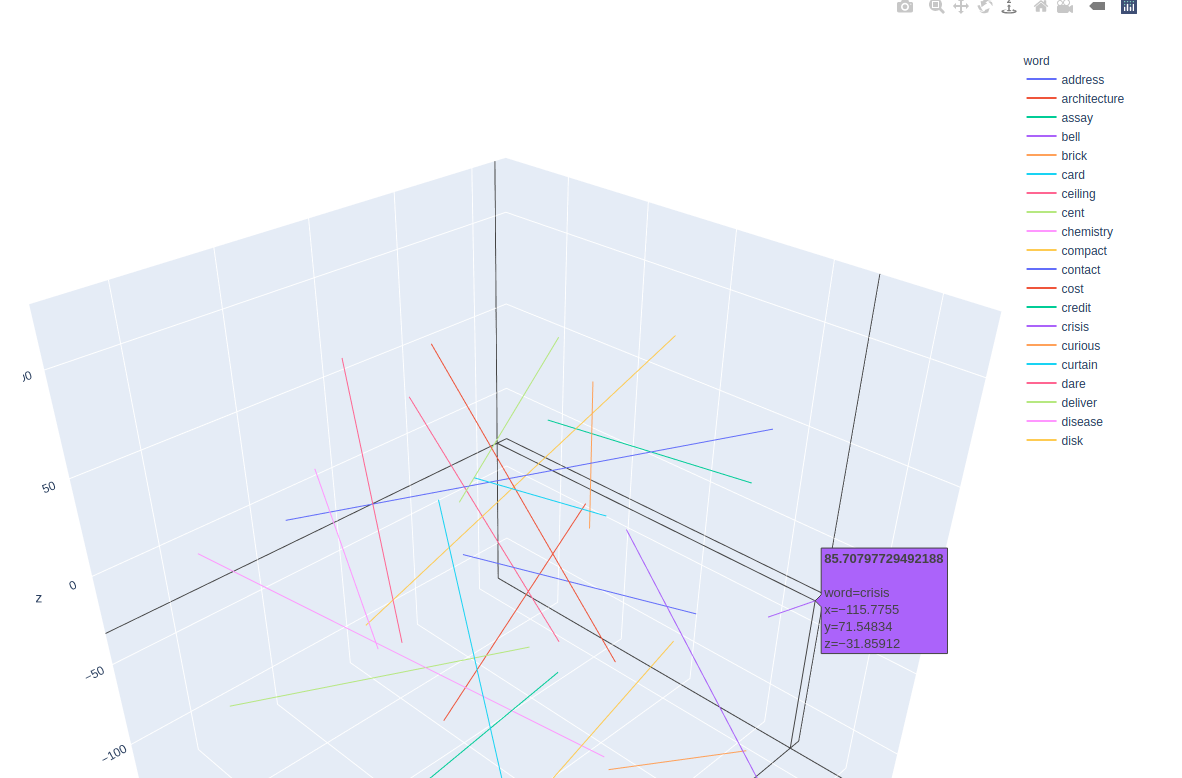}
\caption{Visualization of semantic change using 3D scatter plot where embeddings, that are projected into the 3D scape, are united with a line if they represent the same word }
\label{fig:graphs}
\end{figure}

\begin{figure}[!ht]
\centering
\includegraphics[width=0.8\columnwidth]{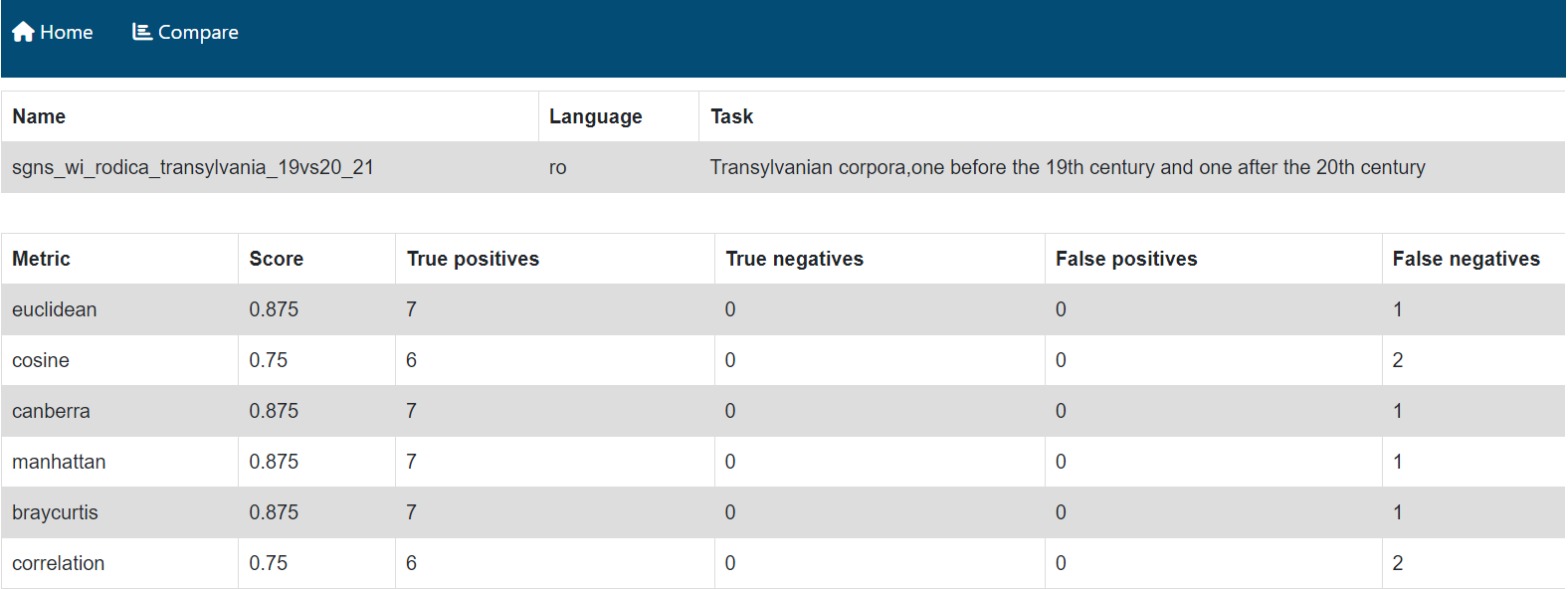}
\caption{Results page presenting the selected model and the obtained results}
\label{fig:table}
\end{figure}

\begin{figure}[!ht]
    \centering
    \subfloat[All metrics selected\label{fig:compare}]{{\includegraphics[width=0.5\columnwidth]{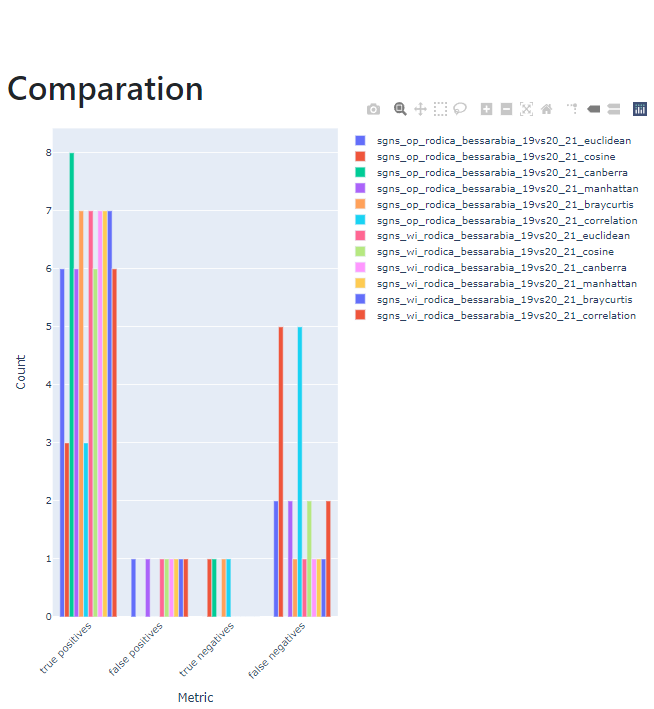}}}
    \hfill    
    \subfloat[Some metrics selected\label{fig:compare_focused}]{{\includegraphics[width=0.5\columnwidth]{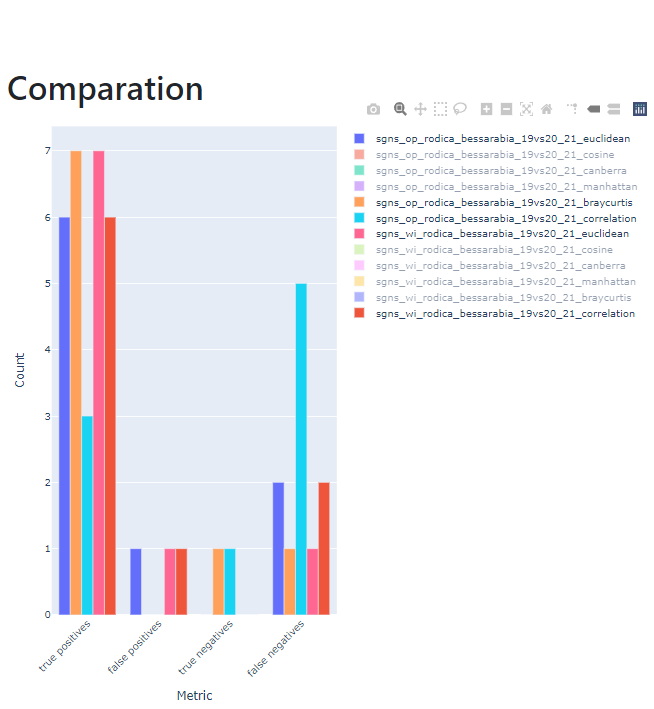}}}
    \caption{Metrics comparison graph}
    \label{fig:comparison}
\end{figure}

\paragraph{Visualization and exploration}
To visualize and explore the results of our models, we created a web application using \href{https://flask.palletsprojects.com/en/2.2.x/}{Flask}.
The application contains the following pages created dynamically using \href{https://palletsprojects.com/p/jinja/}{Jinja2}:
(1) Index: a page that lists the selected models.
(2) Results: a page for visualizing the results of the different models (Figures~\ref{fig:graphs} and~\ref{fig:table}).
(3) Select: a page for selecting models for comparison.
(4) Compare: a page that presents the comparison results for the selected models (Figures~\ref{fig:compare}, and~\ref{fig:compare_focused}).

The projection from the embedding space, which has at least 100 dimensions, to the 3D space is done using the t-distributed stochastic neighbor embedding (t-SNE) method \cite{VanderMaaten2008}.
The t-SNE algorithm is composed of two stages:
\begin{enumerate}
    \item The algorithm creates a probability distribution over pairs of data entities, where very similar entities are assigned a higher probability and dissimilar objects are assigned a lower probability
    \item The algorithm creates a set of points in a low dimensionality space, creates a similar probability distribution and minimizes the Kullback-Leibler divergence between the two distributions concerning the location of points in space
\end{enumerate}

The graphics are created using the \href{https://plotly.com/python/getting-started/}{Plotly} plotting framework.
We choose this framework since the graphics could be directly embedded into the web pages and they would be interactive.
The graphics can be manipulated in a variety of ways (Figure~\ref{fig:compare_focused}), e.g., selecting/deselecting metrics for visualization, selecting/deselecting the model for analysis, etc. 
Furthermore, these graphs can be saved.

\section{Experimental Results}~\label{sec:results}
In this section, we present our datasets and experimental setup, and we analyze the obtained results.

\subsection{Datasets details}

\paragraph{SEMEVAL-CCOHA (SemEval Clean Corpus of Historical American English)}
The first set of experiments is conducted on the SEMEVAL-CCOHA corpra~\cite{Schlechtweg2020}.
We use these corpora to test the different word embeddings and evaluate their performance before moving to the Romanian corpus.

\paragraph{RODICA (ROmanian DIachonic Corpus with Annotations)}
RODICA is a collection of 153 Romanian language new articles collected from the four historical regions of current-day Romania: Wallachia, Transylvania, Moldovia, and Bessarabia.
The news articles where published in the local press, spanning over a time period that covers the half of the 19th century until the early part of the 21th century
Table~\ref{tab:rodica_region_time_count} presents the distribution of texts by region and by date.

\begin{table}[!htbp]
    \centering
    \caption{RODICA documents distribution}
    \label{tab:rodica_region_time_count} 
	\begin{tabular}{ |l|r|r|r|r|}
	    \hline
	    \textbf{Region}& \textbf{19th century}& \textbf{20th century}&\textbf{21st century}& \textbf{Total} \\
	    \hline
	    Bessarabia	&4	&63	&0	&67\\
	    Moldavia	&11	&3	&6	&20\\
	    Transylvania	&37		&0	&7	&44\\
	    Wallachia	&5	&60	&0	&19\\
	    All  &58	&63	&32	&153\\
	    \hline
	\end{tabular}
\end{table}

\subsection{Experimental setup}

For the SEMEVAL-CCOHA corpus, we use the textual data as provided in~\cite{Schlechtweg2020}.
For RODICA, we create 4 corpora for each of the historical regions:
(1) RODICA-BS, the text from the region of Bessarabia;
(2) RODICA-MD, the text from the region of Moldavia;
(3) RODICA-TR, the text from the region of Transylvania;
(4) RODICA-WL, the text from the region of Wallachia.
Before training the word embeddings and to mitigate against the small number of text in the dataset written in the 21st century, we split each dataset into 2 time periods: before and after 1900.
We obtain the following pairs of corpora that are used together in the tests. 

We use The Catalogue of Semantic Shifts~\cite{Zalizniak2012} for Romanina to obtain a list that contains only words that changed their meaning.
We also use Swadesh~\cite{Novotna2007} to create a second list with of words that tend to change their meaning rather slowly in a language at most of their levels, semantic and phonetic, being mainly used to asses when the split between two languages occurred in their ancestry
For these 2 lists, we keep only the words that appear in both time periods for a given region.
We use the evaluation metrics with these lists to determine if we manage to detect correctly semantic changes by determining if the word maintains its ranking,  and thus its meaning, in the list of words created using Swadesh.

For the experiments, we train the following models:
(1) SGNS-OP: the representation model formed with Skip-Gram negative sampling and Orthogonal Procrustes, and
(2) SGNS-WI: the representation model formed with Skip-Gram negative sampling and Word Injection,
(3) ELMO-PREV: an ELMo model trained on the diachronic corpus with text from the first time period;
(4) ELMO-POST: an ELMo model trained on the diachronic corpus with text from the second time period.

\subsection{Results on SEMEVAL-CCOHA}

\textbf{Static word embeddings.}
The first set of experiments uses SGNS-OP and SGNS-WI on the SEMEVAL-CCOHA. 
We observe that both approaches obtain similar results (Table~\ref{tab:sgns_semeval}). 

\begin{table}[!htbp]
    \centering
    \caption{SEMEVAL-CCOHA results (Abbreviations: TP - True positive, TN - True Negative, FP - False Positive, FN - False Negative)}
    \label{tab:sgns_semeval} 
	\begin{tabular}{ |l|l|l|r|r|r|r|  }
		\hline
		\textbf{Model}& \textbf{Metric} & \textbf{Score} & \textbf{TP} & \textbf{TN} & \textbf{FP} & \textbf{FN}\\
		\hline
		\multirow{6}{*}{SGNS-OP}
		&Euclidean   &0.73&39&0&2&12\\
		&Manhattan   &0.73&39&0&2&12\\
		&Canberra   &0.90&48&0&2&3\\
            &Cosine   &0.56&29&1&1&22\\
		&Bray Curtis   &0.86&46&0&2&5\\
		&Correlation   &0.56&29&1&1&22\\
		\hline
		\multirow{6}{*}{SGNS-WI}
		&Euclidean   &0.72&38&0&2&13\\
		&Manhattan   &0.72&38&0&2&13\\
		&Canberra   &0.89&47&0&2&4\\
            &Cosine   &0.57&29&1&1&22\\
		&Bray Curtis   &0.85&45&0&2&6\\
		&Correlation   &0.57&29&1&1&22\\
		\hline
	\end{tabular}
\end{table}

For both models, by intersecting the various metrics, we can observe that Euclidean and Manhattan distance have the highest correlation.
Another strong correlation can be observed between the Consine and The Bray-Cutis distances.
Table~\ref{tab:words_sgns_op_semeval} present the words that undergo semantic change detected by the SGNS-OP model w.r.t. each metric, while Table~\ref{tab:words_sgns_wi_semeval} present the words that undergo semantic change detected by the SGNS-WI model w.r.t. each metric 

\begin{table}[!ht]
    \centering
    \caption{Semantic changes detected using SGNS-OP on SEMEVAL-CCOHA}
    \label{tab:words_sgns_op_semeval} 
 \begin{tabular}{ |l|l|l|l|l|l|  }
		\hline
		\textbf{Euclidean} & \textbf{Manhattan} & \textbf{Canberra} & \textbf{Cosine} & \textbf{Bray-Curtis} & \textbf{Correlation} \\ \hline
justice            & justice            & justice           & lift            & disease              & lift                 \\
federal            & federal            & leaf              & disease         & lift                 & disease              \\
address            & address            & maid              & drug            & drug                 & drug                 \\
maybe              & maybe              & disease           & leaf            & justice              & leaf                 \\
maid               & maid               & lift              & justice         & leaf                 & justice              \\
lift               & lift               & drug              & brick           & maid                 & maid                 \\
leaf               & gain               & domain            & maid            & excuse               & brick                \\
gain               & leaf               & excuse            & disorder        & domain               & disorder             \\
disorder           & drug               & disorder          & excuse          & brick                & excuse               \\
drug               & disease            & cost              & domain          & disorder             & domain               \\ \hline
	\end{tabular}
\end{table}

\begin{table}[!htbp]
    \centering
    \caption{Semantic changes detected using SGNS-WI on SEMEVAL-CCOHA}
    \label{tab:words_sgns_wi_semeval} 
 \begin{tabular}{|l|l|l|l|l|l|}
\hline
\textbf{Euclidean} & \textbf{Manhattan} & \textbf{Canberra} & \textbf{Cosine} & \textbf{Bray-Curtis} & \textbf{Correlation} \\ \hline
honour             & disease            & disease           & disk            & maid                 & disk                 \\
disease            & honour             & honour            & credit          & assay                & credit               \\
deliver            & deliver            & disk              & family          & dining               & family               \\
address            & address            & gallery           & federal         & disk                 & federal              \\
excuse             & excuse             & doubt             & martial         & context              & martial              \\
lip                & lip                & lip               & disorder        & credit               & disease              \\
federal            & federal            & federal           & disease         & architecture         & disorder             \\
gallery            & gallery            & family            & maid            & martial              & maid                 \\
energy             & energy             & cost              & architecture    & family               & honour               \\
family             & family             & extracellular     & honour          & compact              & dining               \\ \hline
\end{tabular}
\end{table}

\textbf{Contextual word embeddings.}
Table~\ref{tab:elmo_semeval} presents the best performing average pairwise distance is APD Cosine followed by Jensen–Shannon divergence for both ELMO-PREV and ELMO-POST on SEMEVAL-CCOHA.

\begin{table}[!ht]
    \centering
    \caption{\textcolor{black}{Results obtained for ELMO-PREV and ELMO-POST on SEMEVAL-CCOHA}}
    \label{tab:elmo_semeval} 

 \begin{tabular}{ |l|l|r|r|r|r|r|  }
		\hline
		\textbf{Model}& \textbf{Metric} & \textbf{Score} & \textbf{TP} & \textbf{TN} & \textbf{FP} & \textbf{FN}\\
		\hline
		\multirow{6}{*}{ELMO-PREV}
		&APD Euclidean   &0.15&8&7&0&83\\
            &APD Manhatten   &0.15&8&7&0&83\\
		&APD Canberra   &0.15&8&7&0&83\\
  		&APD Cosine   &0.78&75&1&6&16\\
		&Jensen–Shannon divergence   &0.55&53&1&6&38\\
		&Cluster Count   &0.52&48&3&4&43\\
		\hline
		\multirow{6}{*}{ELMO-POST}
		&APD Euclidean   &0.20&13&7&0&78\\
            &APD Manhatten   &0.20&13&7&0&78\\
		&APD Canberra   &0.20&13&7&0&78\\
            &APD Cosine   &0.78&75&1&6&16\\
		&Jensen–Shannon divergence   &0.50&48&1&6&43\\
		&Cluster Count   &0.39&34&4&3&57\\
		\hline
	\end{tabular}
\end{table}

The first assumption we made regarding contextualized embeddings is that they can more accurately represent the various nuances in the meaning of a word.
Thus, we choose for experiments some words that were listed in the Catalogue of Semantic Shifts as Polysemy (acquisition of new meaning by a word).
For example, the word \textit{"ceiling"} (with the meaning \textit{"the hard upper part of the inside of the mouth"} in addition to its original sense) is 
detected by both ELMo models as semantic change,  while the SGNS-OP and SGNS-WI models do not detect it. 
This result is most likely a false positive since the word did not acquire this sense in the second corpus.

When ranking the semantic changes (Tables~\ref{tab:words_elmo_prev} and~\ref{tab:words_elmo_post}), we observe that the contextual word embedding models detect several semantic changes that are not listed in the Catalogue of Semantic Shifts, e.g., \textit{orange}. 
At first, we might think this captures the fact that at some point in time, the word for the orange (the fruit) also began to mean the color of the fruit.
But the earliest recorded usage of this word as color comes from 1512 and dates before this year are not included in SEMEVAL-CCOHA.

Analyzing only the words that both models detected that have undergone semantic change, we obtain different results.
We stored the true positives for every metric of both models (Table~\ref{tab:elmo_semeval}).
The best performing metric for both ELMo models is Cosine which also manages to detect all the words from the other metrics.
Furthermore, the lists of semantic changes detected with APD Cosine for ELMO-PREV and ELMO-POST are identical. 
We also observe using the Cluster Count metric that with ELMo we obtain more regular embeddings~\cite{contextual}.

\begin{table}[!ht]
    \centering
    \caption{Semantic changes detected using ELMO-PREV on SEMEVAL-CCOHA}
    \label{tab:words_elmo_prev} 
\begin{tabular}{|l|l|l|l|l|l|l|}
\hline
\textbf{Euclidean} & \textbf{Manhattan} & \textbf{Canberra} & \textbf{Cosine} & \textbf{Bray-Curtis} & \textbf{Correlation}  \\ \hline
disorder           & virtual            & disorder          & virtual         & disorder             & virtual              \\
disk               & strategy           & disk              & strategy        & disk                 & strategy            \\
ceiling            & spine              & ceiling           & spine           & ceiling              & spine                \\
negligence         & replacement        & negligence        & replacement     & negligence           & replacement         \\
upwards            & program            & upwards           & program         & upwards              & program              \\
assay              & participate        & assay             & participate     & prestige             & participate          \\
prestige           & node               & prestige          & node            & assay                & node                 \\
optical            & mirror             & optical           & mirror          & optical              & mirror               \\
virtual            & jet                & virtual           & jet             & virtual              & jet                  \\
strategy           & infection          & strategy          & infection       & strategy             & infection            \\ \hline
\end{tabular}
\end{table}

\begin{table}[!ht]
    \centering
    \caption{Semantic changes detected using ELMO-POST on SEMEVAL-CCOHA}
    \label{tab:words_elmo_post} 
	\begin{tabular}{|l|l|l|l|l|l|l|}
\hline
\textbf{Euclidean} & \textbf{Manhattan} & \textbf{Canberra} & \textbf{Cosine} & \textbf{Bray-Curtis} & \textbf{Correlation}  \\ \hline
disk               & woman              & disorder          & structure       & disk                 & lip                                 \\
disorder           & wine               & disk              & orange          & disorder             & credit                         \\
negligence         & west               & negligence        & mirror          & negligence           & address                          \\
jet                & virus              & ceiling           & disease         & jet                  & bell                            \\
ceiling            & virtual            & jet               & cent            & ceiling              & cost                                \\
node               & vector             & node              & crisis          & node                 & scene                               \\
virtual            & vaccine            & optical           & gain            & virtual              & justice                            \\
prestige           & user               & replacement       & sleep           & prestige             & west                                \\
replacement        & upwards            & prestige          & lift            & replacement          & gain                              \\
upwards            & trap               & virtual           & drug            & upwards              & parent                           \\ \hline
\end{tabular}
\end{table}

Table \ref{tab:elmo_semeval} presents some of the words detected by the contextual word embedding models that undergo semantic change.

\begin{table}[!htbp]
    \centering
    \caption{SEMEVAL-CCOHA: semantic change examples}
    \label{tab:elmo_shift} 
 \begin{tabular}{ |l|p{10cm}| }
		\hline
		\textbf{Word}& \textbf{Change} \\
		\hline
        lift &this is an example of the emergence of a new meaning for a word as a response to the evolution of technology. In the second corpora, the word lift is used in the context \textit{"all this have been reporting to the management by the lift boy"} where the term "lift" refers to the elevator, while it refers to the verb "to lift" for all the other instances as in the context \textit{"the angel shall stand upon the sea and the earth and lift up his hand to heaven"} \\
        \hline
        bell &this word is used in the first corpus with the meanings of "bell"(musical instrument), \textit{"bell of cowslip"(a flower)}, and a surname. In the second corpus the meaning of "bell of cowslip" is not encountered. \\
        \hline
        leaf &In the first corpus leaf is used both in its literal sense as a part of the plant as in the context \textit{"tree with rough leaf"}, and in a metaphoric use in expressions such as \textit{"say the fiend and he shakes like a leaf when cast"} and \textit{"the leaf of the book once"}. In the second corpus, it is used only in its denominative sense \\
        \hline
        mirror &this is most likely a false positive since the newly acquired meaning of the word mirror from the Catalogue of Semantic Shift is the meaning of "a  copy of an internet page to be used if the original is down". \\
        \hline
	\end{tabular}
\end{table}

Although the cosine distance performs better with the contextual embedding models, non-contextualized embeddings performed better.
Our findings are similar to the findings of Basile et al. (2020)~\cite{Basile2020}.
Thus, because of the small size of the RODICA corpus and a very small performance increase for the cosine distance, we choose not to perform experiments with ELMo in the RODICA corpus.

\subsection{Results on RODICA}

For this set of experiments, we train both models on each of the 4 regional datasets. 
We assess the models' performance by looking at how high in the ranking the words that did not change its semantic properties are positioned, i.e., \textit{aici} (\textit{here}) and \textit{acolo} (\textit{there}).
Note: in the following tables, orange marks the word(s) that did not undergo any semantic change.

\paragraph{Bessarabia Region.}
For this dataset, we observe that SGNS-OP performs slightly better than SGNS-WI (Tables~\ref{tab:words_sgns_op_rodica_bs} and~\ref{tab:words_sgns_wi_rodica_bs}).
The word \textit{aici} ranked 5 out of 10 words when using the Cosine and Correlation distances.
Furthermore, we observe a strong correlation (i.e., the ranking lists are similar to identical) among the results obtained with the following distance pairs: $<$ Euclidean, Manhattan $>$, $<$ Cosine, Correlation $>$, and $<$ Canberra, Bray-Curtis $>$.
Table~\ref{tab:bs_shift} discusses some of the semantic changes detected.

\begin{table}[!htbp]
    \centering
    \caption{SGNS-OP: RODICA-BS Semantic changes}
    \label{tab:words_sgns_op_rodica_bs} 
\begin{tabular}{|l|l|l|l|l|l|}
\hline
\textbf{Euclidean}     & \textbf{Manhattan}     & \textbf{Canberra}      & \textbf{Cosine}        & \textbf{Bray-Curtis}   & \textbf{Correlation}   \\ \hline
mare                   & mare                   & grâu                   & grâu                   & grâu                   & grâu                   \\
\cellcolor{orange}aici & \cellcolor{orange}aici & vară                   & vară                   & vară                   & vară                   \\
grâu                   & grâu                   & \cellcolor{orange}aici & mie                    & \cellcolor{orange}aici & mie                    \\
mic                    & mic                    & mare                   & mic                    & mare                   & mic                    \\
vară                   & vară                   & mie                    & \cellcolor{orange}aici & mic                    & \cellcolor{orange}aici \\
mie                    & mie                    & mic                    & cale                   & mie                    & cale                   \\
inimă                  & inimă                  & inimă                  & inimă                  & inimă                  & inimă                  \\
cale                   & cale                   & cale                   & parte                  & cale                   & parte                  \\
parte                  & parte                  & parte                  & mare                   & parte                  & mare                   \\ \hline
\end{tabular}
\end{table}

\begin{table}[!ht]
    \centering
    \caption{SGNS-WI: RODICA-BS Semantic changes}
    \label{tab:words_sgns_wi_rodica_bs} 
\begin{tabular}{|l|l|l|l|l|l|}
\hline
\textbf{Euclidean}     & \textbf{Manhattan}     & \textbf{Canberra}      & \textbf{Cosine}        & \textbf{Bray-Curtis}   & \textbf{Correlation}   \\ \hline
mare                   & mare                   & \cellcolor{orange}aici & mare                   & \cellcolor{orange}aici & mare                   \\
\cellcolor{orange}aici & \cellcolor{orange}aici & mare                   & \cellcolor{orange}aici & mare                   & \cellcolor{orange}aici \\
cale                   & cale                   & cale                   & cale                   & cale                   & cale                   \\
parte                  & parte                  & vară                   & parte                  & vară                   & parte                  \\
inimă                  & inimă                  & inimă                  & inimă                  & grâu                   & inimă                  \\
vară                   & vară                   & parte                  & mie                    & inimă                  & mie                    \\
grâu                   & grâu                   & grâu                   & vară                   & parte                  & vară                   \\
mie                    & mie                    & mie                    & mic                    & mie                    & mic                    \\
mic                    & mic                    & mic                    & grâu                   & mic                    & grâu                   \\ \hline
\end{tabular}
\end{table}

\begin{table}[!ht]
    \centering
    \caption{RODICA-BS: semantic change examples}
    \label{tab:bs_shift} 
 \begin{tabular}{ |l|p{10cm}| }
		\hline
		\textbf{Word}& \textbf{Change} \\
		\hline
        cale & this word is used in the before 1900 corpus with the sense "to find adequate" in the context \textit{"Prea Sfântul Sinod nu va găsi cu cale a-i da arătată Eparhie"} (The Holy Synod won't find it adequate to give him an Eparchy), in the after 1900 corpus is used with several meanings, some of them being "method" \textit{"pe cale de economii aspre"} (with harsh economic methods) \\
        \hline
        vară & this is most likely a false positive given by a difference in the number of usages across corpora\\
        \hline
	\end{tabular}
\end{table}

\paragraph{Moldavia Region.} 
The Moldavia corpora are somewhat balanced being made out of 11 texts for the period before 1900 and 9 texts for the period after 1900.
The best results are obtained using the SGNS-OP model together with the Cosine and Correlation distances (Tables~\ref{tab:words_sgns_op_rodica_md} and~\ref{tab:words_sgns_wi_rodica_md}), which also obtain the same rankings.
Table~\ref{tab:md_shift} discusses some of the detected semantic changes.

\begin{table}[!htbp]
    \centering
    \caption{SGNS-OP: RODICA-MD Semantic changes}
    \label{tab:words_sgns_op_rodica_md} 
\begin{tabular}{|l|l|l|l|l|l|}
\hline
\textbf{Euclidean}      & \textbf{Manhattan}      & \textbf{Canberra}       & \textbf{Cosine}         & \textbf{Bray-Curtis}    & \textbf{Correlation}    \\ \hline
parte                   & parte                   & grâu                    & grâu                    & grâu                    & grâu                    \\
\cellcolor{orange}aici  & \cellcolor{orange}aici  & \cellcolor{orange}aici  & drum                    & \cellcolor{orange}aici  & drum                    \\
mare                    & mare                    & țara                    & mie                     & țara                    & mie                     \\
țara                    & țara                    & parte                   & deschis                 & drum                    & deschis                 \\
\cellcolor{orange}acolo & \cellcolor{orange}acolo & drum                    & lume                    & parte                   & lume                    \\
drum                    & drum                    & mare                    & \cellcolor{orange}aici  & mare                    & \cellcolor{orange}aici  \\
grâu                    & grâu                    & mie                     & țara                    & mie                     & țara                    \\
mie                     & mie                     & \cellcolor{orange}acolo & \cellcolor{orange}acolo & \cellcolor{orange}acolo & \cellcolor{orange}acolo \\
deschis                 & deschis                 & deschis                 & mare                    & deschis                 & mare                    \\
lume                    & lume                    & lume                    & parte                   & lume                    & parte                   \\ \hline
\end{tabular}
\end{table}

\begin{table}[!ht]
    \centering
    \caption{SGNS-WI: RODICA-BS Semantic changes}
    \label{tab:words_sgns_wi_rodica_md} 
\begin{tabular}{|l|l|l|l|l|l|}
\hline
\textbf{Euclidean}      & \textbf{Manhattan}      & \textbf{Canberra}       & \textbf{Cosine}         & \textbf{Bray-Curtis}    & \textbf{Correlation}    \\ \hline
parte                   & parte                   & \cellcolor{orange}acolo & parte                   & \cellcolor{orange}acolo & parte                   \\
\cellcolor{orange}aici  & \cellcolor{orange}aici  & \cellcolor{orange}aici  & \cellcolor{orange}aici  & parte                   & \cellcolor{orange}aici  \\
\cellcolor{orange}acolo & \cellcolor{orange}acolo & parte                   & \cellcolor{orange}acolo & \cellcolor{orange}aici  & \cellcolor{orange}acolo \\
mare                    & mare                    & cale                    & mare                    & cale                    & mare                    \\
cale                    & cale                    & mare                    & greu                    & mare                    & greu                    \\
față                    & față                    & greu                    & cale                    & față                    & cale                    \\
greu                    & greu                    & întreg                  & întreg                  & întreg                  & întreg                  \\
întreg                  & întreg                  & față                    & față                    & greu                    & față                    \\ \hline
\end{tabular}
\end{table}

\begin{table}[!ht]
    \centering
    \caption{RODICA-MD: semantic change examples}
    \label{tab:md_shift} 
 \begin{tabular}{ |l|p{10cm}| }
		\hline
		\textbf{Word}& \textbf{Change} \\
		\hline
        grâu & this words appear to change due to a difference in the number of its usage across corpora, having four usages in the first corpus and only one in the second corpus\\
        \hline
        deschis & this word is used in with the sense "open" in the corpus with texts after 1900, while  is used in a more figurative manner in the corpus with texts before 1900: \textit{"pocnetul deschis"} (the loud noise)\\
        \hline
	\end{tabular}
\end{table}

\paragraph{Transylvania Region.}
We observe that SGNS-WI does not manage to correctly identify semantic change as the words that do not change their sense are ranked last (Table~\ref{tab:words_sgns_op_rodica_tr}).
In contrast, SNGS-OP obtains good results managing to determine correctly semantic changes (Table~\ref{tab:words_sgns_wi_rodica_tr}).
A particularity of this corpora is the high occurrence of non-standard spelling in the before 1900 corpus, containing diacritics originating from an earlier form of the Romanian Latin Alphabet and non-standard phonetic realizations, e.g., \textit{"dobendí"} instead of the correct standard Contemporary Romanian word \textit{"dobândi"}.
Table~\ref{tab:tr_shift} discusses some of the detected semantic changes.

\begin{table}[!ht]
    \centering
    \caption{SGNS-OP: RODICA-TR Semantic changes}
    \label{tab:words_sgns_op_rodica_tr} 
\begin{tabular}{|l|l|l|l|l|l|}
\hline
\textbf{Euclidean}      & \textbf{Manhattan}      & \textbf{Canberra}       & \textbf{Cosine}         & \textbf{Bray-Curtis}    & \textbf{Correlation}    \\ \hline
mare                    & mare                    & față                    & întreg                  & întreg                  & întreg                  \\
față                    & față                    & întreg                  & față                    & față                    & față                    \\
întreg                  & întreg                  & greu                    & cale                    & greu                    & cale                    \\
greu                    & greu                    & cale                    & greu                    & cale                    & greu                    \\
cale                    & cale                    & mare                    & mare                    & mare                    & mare                    \\
parte                   & parte                   & \cellcolor{orange}acolo & \cellcolor{orange}aici  & \cellcolor{orange}acolo & \cellcolor{orange}aici  \\
\cellcolor{orange}acolo & \cellcolor{orange}acolo & parte                   & \cellcolor{orange}acolo & parte                   & \cellcolor{orange}acolo \\
\cellcolor{orange}aici  & \cellcolor{orange}aici  & \cellcolor{orange}aici  & parte                   & \cellcolor{orange}aici  & parte                   \\ \hline
\end{tabular}
\end{table}

\begin{table}[!ht]
    \centering
    \caption{SGNS-WI: RODICA-TR Semantic changes}
    \label{tab:words_sgns_wi_rodica_tr} 
\begin{tabular}{|l|l|l|l|l|l|}
\hline
\textbf{Euclidean}      & \textbf{Manhattan}      & \textbf{Canberra}       & \textbf{Cosine}         & \textbf{Bray-Curtis}    & \textbf{Correlation}    \\ \hline
parte                   & parte                   & \cellcolor{orange}acolo & parte                   & \cellcolor{orange}acolo & parte                   \\
\cellcolor{orange}aici  & \cellcolor{orange}aici  & \cellcolor{orange}aici  & \cellcolor{orange}aici  & parte                   & \cellcolor{orange}aici  \\
\cellcolor{orange}acolo & \cellcolor{orange}acolo & parte                   & \cellcolor{orange}acolo & \cellcolor{orange}aici  & \cellcolor{orange}acolo \\
mare                    & mare                    & cale                    & mare                    & cale                    & mare                    \\
cale                    & cale                    & mare                    & greu                    & mare                    & greu                    \\
față                    & față                    & greu                    & cale                    & față                    & cale                    \\
greu                    & greu                    & întreg                  & întreg                  & întreg                  & întreg                  \\
întreg                  & întreg                  & față                    & față                    & greu                    & față                    \\ \hline
\end{tabular}
\end{table}

\begin{table}[!ht] 
    \centering
    \caption{RODICA-TR: semantic change examples}
    \label{tab:tr_shift} 
 \begin{tabular}{ |l|p{10cm}| }
		\hline
		\textbf{Word}& \textbf{Change} \\
		\hline
        greu & in the before 1900 corpus the word is used as an adjective in contexts such as \textit{"drumulu acelu greu alu publicitâtii"} (that hard way of publicity), while in the after 1900 corpus is used more as an adverb in contexts such as \textit{"este greu de crezut"}(it's hardly believable) \\
        \hline
        vară & this is most likely a false positive given by a difference in the number of its usage across the corpora\\
        \hline
	\end{tabular}
\end{table}

\paragraph{Wallachia Region.}
Both models do not obtain satisfactory results on the RODICA-WL.
The RODICA-WL corpus with texts after the 1900 does not contain any of the  Swadesh words. Thus, we cannot use the same ranking method for our analysis.
Thus, we test for semantic change the following 3 words: \textit{lună} (moon), \textit{țară} (county), and \textit{parte} (part). 
We observe that the models determine that the word \textit{ "parte"} has the highest semantic change regardless of the metric employed (Table~\ref{tab:words_sgns_rodica_wl}).
Table~\ref{tab:wl_shift} presents some of the detected semantic changes.

\begin{table}[!ht]
    \centering
    \caption{RODICA-WL Semantic changes}
    \label{tab:words_sgns_rodica_wl}
 \begin{tabular}{ |l|l|l|l|l| }
		\hline
		\textbf{Model} &\textbf{Metric}& \textbf{Word 1} & \textbf{Word 2} & \textbf{Word 3} \\
		\hline
            \multirow{6}{*}{\textbf{SGNS-WI}}
            &\textbf{Euclidean}& parte& lună& țară \\
            &\textbf{Manhattan} & parte& lună& țară \\
            &\textbf{Cosine} & parte& lună& țară \\
            &\textbf{Canberra} & parte& lună& țară \\
            &\textbf{Bray-Curtis} & parte& lună& țară \\
            &\textbf{Correlation} &parte& lună& țară \\
            \hline
            \multirow{6}{*}{\textbf{SGNS-OP}}
            &\textbf{Euclidean}& parte& țară& lună \\
            &\textbf{Manhattan} & parte& țară& lună \\
            &\textbf{Cosine} & parte& lună& țară \\
            &\textbf{Canberra} & parte& țară& lună \\
            &\textbf{Bray-Curtis} & parte& țară& lună \\
            &\textbf{Correlation} & parte& lună& țară \\
		\hline
	\end{tabular}
\end{table}

\begin{table}[!ht] 
    \centering
    \caption{RODICA-WL: semantic change examples} 
    \label{tab:wl_shift}
 \begin{tabular}{ |l|p{10cm}| }
		\hline
		\textbf{Word}& \textbf{Change} \\
		\hline
        parte & in the after 1900 corpus is used as "a part of a country" as in the context \textit{"în această parte de țară"}, while in the before 1900 corpus it is used more with the sense of "part of something", e.g., \textit{parte a lucrării pentru articolul viitor} (part of the work for the future article) , or as "to take part in", e.g.,  \textit{boierii luau parte la serviciul divin} (the boyars took part in the divine service)\\
        \hline
        lună & in both corpora the word means \textit{"month"}, as we can see in contexts like \textit{"aproape o lună de dile"} in the second corpus and contexts such as \textit{"Apare de 2 ori pe lună"}\\
        \hline
        țară & this is again a word whose sense did not change, being used in both corpora with the sense of \textit{country} \\
        \hline
	\end{tabular}
\end{table}

\section{Discussions}~\label{sec:discussions}

The cosine distance has the overall worst performance on the SEMEVAL-CCOHA corpus regardless of the model, while Canberra has the best performance in detecting semantic change (Table~\ref{tab:sgns_semeval}).
When considering the same distance metric, both models perform similarly.
We can conclude that the vector space model, i.e., Orthogonal Procrustes and Word Injection, have very little influence over the final word representation for the task of semantic change.

Assessing semantic change for the Romanian Corpora is a non-trivial task since Romanian is usually a low-resource language.
Moreover, we can observe that the Basarabian corpus contains bilingual texts.
Words and expressions from the Russian language appear on many occasions directly in the texts written with latin script, e.g., \textit{za dva leia} which is the Russian phrases \textit{за два лея} (\textit{for two lei}).
These occurrences tend to be rare enough and the performances of the model do not seem to be affected in any significant measure.

Spelling reforms and slight phonetic changes can also affect the performance of the models and are non-trivial problems to overcome when dealing with historic corpora.
An example would be that the Romanian word \textit{religie} could be found in the archive with various spellings corresponding to archaic phonemic realizations such as  \textit{religiune} or \textit{relighie}.
Furthermore, named entities for places also suffer from spelling changes or different naming depending on the region.
This leads to the same concept being treated as two separate ones.
For example, \textit{Elisabethgrad} is the name used in the Transylvania corpus for the city \textit{Kropyvnytskyi} or the city of \textit{Kyiv} appears with the outdated spelling \textit{Kiew}.

\section{Conclusions}~\label{sec:conclusions}

In this paper, we propose a new architecture for semantic change detection on low-resource languages.
For this architecture, we design three strategies to train our word embeddings. 
The first two strategies, i.e., SGNS-OP and SGNS-WI, use static word embeddings. 
The third strategy uses ELMo, a contextual word embedding model.
The proposed static and contextual models are trained on the SEMEVAL-CCOHA English dataset. 
The obtained results act as a baseline for choosing the models to train the Romanian language dataset RODICA.

The experimental results on the English corpus show that the two static word embedding models performs very similarly, while the contextual word embedding model performs poorly in comparison.
For the RODICA corpus, we train the static word embeddings and obtain good results for detecting semantic change by using the Word Injection vector space representation.

In future work, we aim to find better ways to handle historical corpora by testing more models and architectures that could accommodate, especially for low-resource languages, the various diachronic differences in orthography.
We will also consider other embedding techniques such as FastText~\cite{Bojanowski2017}, GloVe~\cite{Pennington2014}, or Hyperbolic Embeddings~\cite{Nickel2017} for the non-contextualized embeddings and BERT~\cite{Devlin2019} for contextualized embeddings.
Also, we will research additional metrics that make better use of contextualized embeddings (e.g., ELMo and BERT).

\bibliographystyle{plainnat}  
\bibliography{main}

\end{document}